\documentclass[letterpaper]{article} 
\usepackage{aaai2026}  
\usepackage{times}  
\usepackage{helvet}  
\usepackage{courier}  
\usepackage[hyphens]{url}  
\usepackage{graphicx} 
\usepackage{natbib}  
\usepackage{caption} 
\usepackage{algorithm}
\usepackage{algorithmic}
\usepackage{tcolorbox}
\usepackage{amsmath}
\usepackage{amsfonts}
\usepackage{multirow}
\usepackage{booktabs}
\usepackage{tablefootnote}
\usepackage{subcaption}
\usepackage[symbol]{footmisc}
\usepackage{placeins}

\definecolor{lightgray}{gray}{0.95}
\usepackage{newfloat}
\usepackage{listings}
\usepackage{docmute}
\DeclareCaptionStyle{ruled}{labelfont=normalfont,labelsep=colon,strut=off} 
\floatstyle{ruled}
\newfloat{listing}{tb}{lst}{}
\floatname{listing}{Listing}
\title{Temporal Cluster Assignment for Efficient Real-Time Video Segmentation}
\author{
    \\
    Ka-Wai Yung\textsuperscript{\rm 1,2}\equalcontrib\thanks{Work done when employed by Medtronic.},
    Felix J. S. Bragman\textsuperscript{\rm 2}\equalcontrib\footnotemark[2],
    Jialang Xu\textsuperscript{\rm 1,2}\equalcontrib\footnotemark[2],
    Imanol Luengo\textsuperscript{\rm 2},
    Danail Stoyanov\textsuperscript{\rm 1,2}, \\
    Evangelos B. Mazomenos\textsuperscript{\rm 1}
}
\affiliations{
    \textsuperscript{\rm 1}UCL Hawkes Institute, University College London, London, UK \\
    \textsuperscript{\rm 2}Medtronic, London, UK\\
}

\begin{document}

\maketitle

\begin{abstract}

Vision Transformers have substantially advanced the capabilities of segmentation models across both image and video domains. Among them, the Swin Transformer stands out for its ability to capture hierarchical, multi-scale representations, making it a popular backbone for segmentation in videos. However, despite its window-attention scheme, it still incurs a high computational cost, especially in larger variants commonly used for dense prediction in videos. This remains a major bottleneck for real-time, resource-constrained applications. Whilst token reduction methods have been proposed to alleviate this, the window-based attention mechanism of Swin requires a fixed number of tokens per window, limiting the applicability of conventional pruning techniques.  Meanwhile, training-free token clustering approaches have shown promise in image segmentation while maintaining window consistency. Nevertheless, they fail to exploit temporal redundancy, missing a key opportunity to further optimize video segmentation performance.

We introduce Temporal Cluster Assignment (TCA), a lightweight and effective, fine-tuning-free strategy that enhances token clustering by leveraging temporal coherence across frames. Instead of indiscriminately dropping redundant tokens, TCA refines token clusters using temporal correlations, thereby retaining fine-grained details while significantly reducing computation. Extensive evaluations on YouTube-VIS 2019, YouTube-VIS 2021, OVIS, and a private surgical video dataset show that TCA consistently boosts the accuracy–speed trade-off of existing clustering-based methods. Our results demonstrate that TCA generalizes competently across both natural and domain-specific videos. 
\end{abstract}


\section{Introduction}

\label{sec:intro}

\begin{figure}[t]
\begin{center}
   \includegraphics[width=\linewidth]{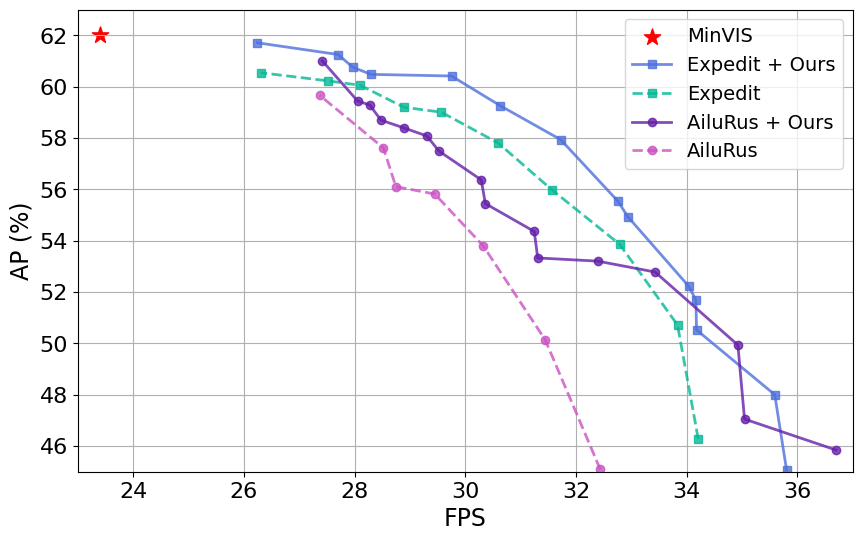}
\end{center}
   \caption{Pareto frontier illustrating the optimal AP–FPS curve for existing token clustering methods and the impact of incorporating our Temporal Cluster Assignment on the YouTube-VIS 2019 dataset. The addition of our method consistently improves the accuracy–speed trade-off across all settings.}
\label{ytvis19_pareto}
\end{figure}

Recent developments in Transformers have driven substantial progress in video segmentation tasks, with impressive results across Video Instance Segmentation (VIS) \cite{wu2022seqformer, cheng2021mask2former, huang2022minvis, heo2022vita, heo2023generalized}, Video Object Segmentation (VOS) \cite{athar2023tarvis, karim2023med, wu2023scalable, yuan2023isomer, luo2023soc}, and Video Semantic Segmentation (VSS) \cite{li2022video,li2023tube, guo2024vanishing,li2024univs}. However, to reach state-of-the-art (SOTA) results, these models typically rely on large, computationally intensive Transformer backbones, which limits their deployment in real-time, resource-constrained environments.

To address this, token reduction techniques have emerged as a solution to improve efficiency \cite{rao2021dynamicvit, meng2022adavit, fayyaz2022adaptive, liang2022evit,wei2023joint,long2023beyond,wang2024zero}. While effective for classification, such methods degrade segmentation performance by discarding critical spatial information. Segmentation-specific approaches, DToP \cite{tang2023dynamic} and SViT \cite{liu2024revisiting}, mitigate this but require costly fine-tuning, hindering generalization and practicality.

In practice, many video segmentation systems operate in real-time and adopt the Swin Transformer \cite{liu2021swin} as the backbone, given its ability to capture multi-scale features crucial for segmenting objects at varying scales \cite{cheng2021mask2former, huang2022minvis, heo2022vita, heo2023generalized}. However, Swin’s window-based attention requires a fixed number of tokens per window, making it incompatible with current segmentation-specific token reduction methods, which drop tokens arbitrarily \cite{tang2023dynamic, liu2024revisiting}. These approaches rely on fine-tuning to determine pruning locations and result in inconsistent token distributions across windows.  Meanwhile, video-specific methods like Video Token Merging \cite{lee2024video} and Object-Centric Token Merging \cite{kahatapitiya2024object} achieve temporal efficiency by merging tokens across frames. Yet, this requires access to future frames during online inference, making them unsuitable.

Token clustering provides a compelling foundation for efficient token reduction. Recent work \cite{liang2022expediting, li2023ailurus} has shown that clustering tokens at intermediate layers in a fine-tuning-free, non-parametric manner can significantly accelerate dense image segmentation while remaining compatible with window-based attention.  Nevertheless, these methods operate frame-wise and fail to exploit temporal dependencies intrinsic to video data \cite{zhang2012slow, tong2022videomae}. Evidently, this is a missed opportunity that could mitigate the degradation observed at high compression ratios due to excessive information loss.

Motivated by the potential to augment existing token clustering by exploiting temporal correlations across video frames, we propose Temporal Cluster Assignment (TCA), a plug-and-play, fine-tuning-free extension of token clustering for online video segmentation.  TCA incorporates temporal coherence by refining token clusters using information from preceding frames. This temporal guidance enables more aggressive compression than frame-wise clustering while preserving spatial fidelity, critical to segmentation accuracy.  In contrast to approaches like MaskVD \cite{sarkar2024maskvd} and Eventful Transformer \cite{dutson2023eventful}, which discard tokens independently across frames, TCA reuses tokens from previous frames to refine clusters in subsequent ones.  This temporal refinement mechanism allows non-reference frames to be compressed more heavily without substantial accuracy loss, achieving a superior trade-off between computational efficiency and segmentation quality.

We validate TCA across three standard VIS benchmarks: YouTube-VIS 2019, YouTube-VIS 2021, and OVIS—and demonstrate its generalization on a surgical VSS dataset.  Results show that integrating TCA into existing token clustering methods consistently improves the trade-off between segmentation accuracy and computational efficiency. Our key contributions are:

\begin{itemize}
    \item A novel extension of token clustering to video segmentation. We introduce Temporal Cluster Assignment (TCA), which leverages temporal coherence across frames to enhance the accuracy–speed trade-off.  To the best of our knowledge, this is the first exploration of clustering-based token reduction for video segmentation, with demonstrated applicability to both natural and domain-specific settings, such as surgical video analysis.
    \item A fine-tuning-free, plug-and-play module. TCA is compatible with existing clustering methods and requires no fine-tuning, making it broadly applicable and easy to integrate into online video segmentation pipelines.
    \item In comprehensive experiments on YouTube-VIS 2019, YouTube-VIS 2021, OVIS, and a surgical dataset, TCA consistently improves the efficiency-accuracy trade-off of SOTA clustering methods.
\end{itemize}

\section{Related Works}
\label{sec:related_works}

\subsection{Transformers for Video Segmentation}
Vision Transformers have rapidly become the core architecture for dense video segmentation, with many adopting Swin as the model backbone.
Mask2Former-VIS \cite{cheng2021mask2former} extends the original Mask2Former \cite{Cheng_2022_CVPR}, designed for image segmentation, by incorporating 3D spatio-temporal attention for video segmentation.
MinVIS \cite{huang2022minvis} proposes a minimalist Transformer-based VIS pipeline, reducing computational overhead by leveraging a lightweight spatiotemporal association mechanism and minimal instance representations.
VITA \cite{heo2022vita} builds on an off-the-shelf image detector to extract per-frame object tokens, distilling object-specific contexts into object tokens and thereby enabling video-level understanding without relying on spatio-temporal features.
GenVIS \cite{heo2023generalized} extends VITA by introducing a novel training strategy that emphasizes inter-clip relationships through a query-based training pipeline with a unique target label assignment. It also incorporates a memory mechanism to retain information from previous frames, improving long-term object association. 

\begin{figure*}[t]
\begin{center}
   \includegraphics[width=\linewidth]{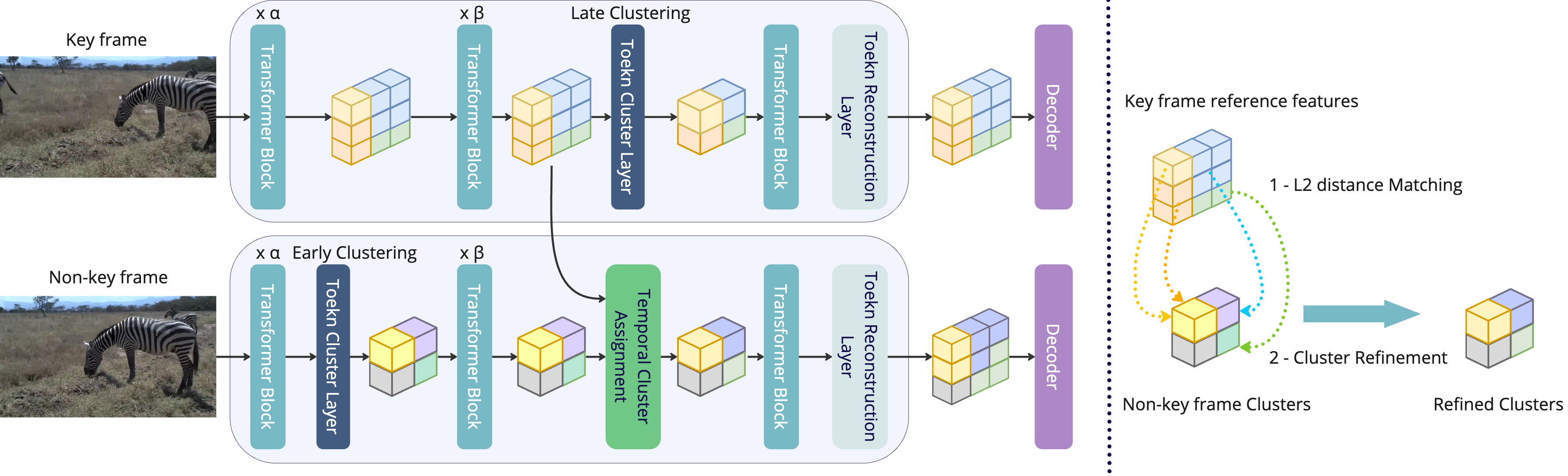}
\end{center}
   \caption{(Left) Overview of our framework. Temporal Cluster Assignment (TCA) leverages temporal correlation by delaying clustering in a reference (key) frame to obtain strong reference tokens, which are then used to refine weaker, early-clustered tokens in subsequent (non-key) frames. (Right) Schematic of the Token Assignment and Refinement process. L2 distance matching is performed to assign each reference token to its closest clustered token. Based on this assignment, clustered tokens are refined using the corresponding reference token through one of the proposed refinement strategies.}
\label{method}
\end{figure*}

\subsection{Efficient Vision Transformers}

Research on efficient Transformers has largely focused on image classification \cite{yin2022vit, fayyaz2022adaptive,meng2022adavit,kong2022spvit,haurum2023tokens,long2023beyond,wei2023joint,wang2024zero}. Methods such as DynamicViT \cite{rao2021dynamicvit} and EViT \cite{liang2022evit}  improve efficiency by pruning tokens deemed less relevant for classification.   ToMe \cite{bolya2022token} and related approaches \cite{kim2024token, norouzi2024algm, song2024moviechat} instead merge semantically similar tokens via bipartite matching, reducing token count while preserving key information.
For image segmentation, DToP \cite{tang2023dynamic} introduces early exits for easy-to-predict tokens, and SViT \cite{liu2024revisiting} selectively processes token subsets using lightweight layers. However, both approaches rely on fine-tuning to determine pruning locations, resulting in inconsistent token distributions across window patches and limited compatibility with window-based attention.
In contrast, Expedit \cite{liang2022expediting} and AiluRus \cite{li2023ailurus} are training-free and window-compatible, using token clustering at intermediate layers followed by full token reconstruction before decoding. Despite their advantages, they are not designed for the video domain and thus overlook both temporal redundancies and dependencies across video frames—an underused signal especially valuable when seeking high compression for aggressive latency optimization.

Video-specific methods like Video Token Merging \cite{lee2024video} and Object-Centric Token Merging \cite{kahatapitiya2024object} address temporal redundancy by comparing and merging tokens across frames. Yet, these techniques assume access to multiple frames, which is common in offline video processing, and thus require future frames at online setting. This makes them unsuitable for online or streaming scenario, where only one frame is available at a time.

\section{Methodology}

\subsection{Preliminaries}
Swin Transformer adopts a hierarchical design by partitioning an input image into non-overlapping patches, which are embedded into token vectors. Instead of relying on global attention across all tokens as in Vision Transformer (ViT), Swin employs a window-based self-attention mechanism (W-MSA) that restricts computation to a local window. At each stage, patch merging operations reduce the spatial resolution while increasing the channel dimension, forming a pyramid-like hierarchy of feature representations. To enable cross-window information exchange, Swin interleaves shifted window configurations so that tokens from adjacent windows can attend to each other in subsequent blocks. 

With its hierarchical architecture, Swin generates multi-scale feature representations capable of capturing both fine-grained and high-level features at different scales, making it effective for segmentation tasks.


\subsection{Temporal Cluster Assignment}

\begin{figure}[t]
\begin{center}
   \includegraphics[width=\linewidth]{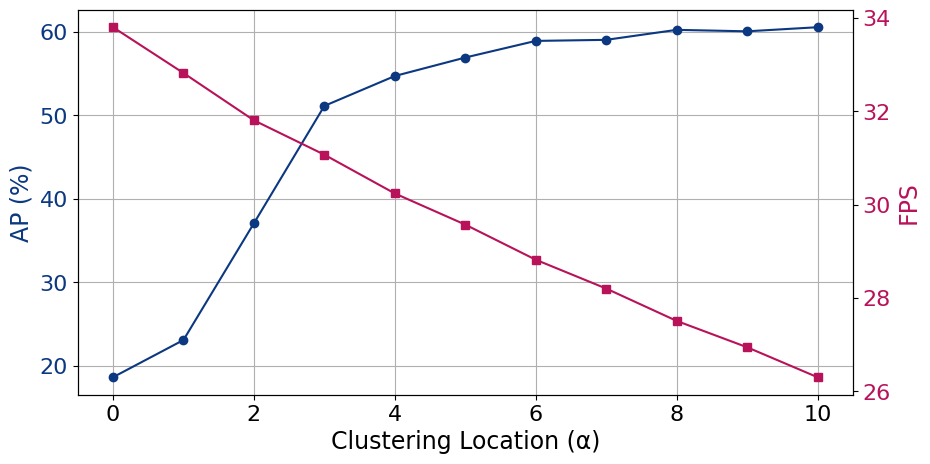}
\end{center}
   \caption{Trade-off between AP and speed across different clustering locations $\alpha$. Earlier clustering yields higher FPS but comes at the cost of performance degradation.}
\label{ap_speed_tradeoff}
\end{figure}


Existing frame-based token clustering methods---Expedit and AiluRus~\cite{liang2022expediting,li2023ailurus}---follow a “high–low–high” resolution pipeline. Given an initial set of $M$ image tokens, at a user-defined intermediate backbone layer $\alpha$, these tokens are clustered into a reduced set of tokens, determined by the cluster size $N$. At a later layer, the clustered tokens are expanded back to their original size $M$ before being fed into the decoder. Since fewer tokens are processed in intermediate layers after clustering, this approach leads to computational speedup. As illustrated in Fig.~\ref{ap_speed_tradeoff}, $\alpha$ introduces a trade-off between speed and performance. Clustering earlier (lower $\alpha$) results in higher speedups but degrade more rapidly. In Swin specifically, token clustering is applied individually within each window due to constraints of window attention, rather than over all tokens globally. As the token cluster size decreases, more tokens within the same window are combined, leading to a loss of fine details and a faster performance drop compared to ViT, where clustering is performed globally.

A straightforward approach to exploit temporal redundancy in videos is to drop similar tokens across consecutive frames. However, Swin requires the same number of tokens per window, making it impossible to simply drop tokens arbitrarily while maintaining dimensional consistency. Whilst MaskVD \cite{sarkar2024maskvd} addresses this by scattering tokens back to their original dimensions before window attention, this operation induces a significant latency overhead.

In contrast, we utilize the temporal correlation and propose TCA. As illustrated in Fig.~\ref{method}, we first process the initial (reference) frame until a later layer ($\alpha + \beta$), allowing tokens to remain unclustered for a longer period. These tokens are then stored before clustering as the reference tokens, denoted as $x_{\text{ref}} \in \mathbb{R}^{K \times M \times L}$, where $K$ is the number of windows, $M$ is the number of tokens per window, and $L$ is the token dimension. For subsequent non-keyframes, we perform earlier clustering $x \in \mathbb{R}^{K \times M \times L} \mapsto \mathbb{R}^{K \times N \times L}$ at layer $\alpha$ to obtain $x_{\text{cluster}}$, where $N$ is the number of clusters per window, with $N \ll M$. We then introduce TCA at layer ($\alpha + \beta$), refining the clustered tokens using the stored reference tokens. To account for feature drifting, a keyframe is selected at a defined interval $f_{\text{max}}$, updating the reference tokens $x_{\text{ref}}$ accordingly. This allows TCA to retain key visual details from the reference frame while still benefiting from early clustering on non-keyframes, striking an optimal balance between performance and efficiency.

To perform cluster refinement, given $x_{\text{ref}}$ and the current frame's $x_{\text{cluster}}$, we first compute the assignment matrix $j^* \in \mathbb{R}^{K \times M}$ through L2 distance matching, as defined in Eq.~\ref{eq:token_matching}. 
\begin{equation}
    j^* = \arg\min_{j} \| x_{\text{ref}, i} - x_{\text{cluster}, j} \|_2, \quad \forall i \in \{1, \dots, M\}
    \label{eq:token_matching}
\end{equation}
where $i$ indexes the tokens in $x_{\text{ref}}$, iterating over all $M$ reference tokens, and $j^*$ denotes the closest match between a reference token and a clustered token.

Using $j^*$, we refine $x_{\text{cluster}}$ by incorporating information from $x_{\text{ref}}$. We explore three different refinement methods:

\paragraph{1) Cluster-Guided Averaging (CGA):} Each $x_{\text{cluster},j}$ is refined by averaging itself with all $x_{\text{ref},i }$ assigned to it:
\begin{equation}
\begin{aligned}
    x_{\text{cluster}, j}^{\text{refined}} &= 
    \frac{1}{1 + |\mathcal{R}_j|} 
    \left( x_{\text{cluster}, j} + \sum_{i \in \mathcal{R}_j} x_{\text{ref}, i} \right)
\end{aligned}
    \label{eq:averaging}
\end{equation}

\paragraph{2) Reference-Based Substitution (RBS):} Each $x_{\text{cluster},j}$ is replaced by the average of all $x_{\text{ref},i }$ assigned to it:
\begin{equation}
    x_{\text{cluster}, j}^{\text{refined}} = \frac{1}{|\mathcal{R}_j|} \sum_{i \in \mathcal{R}_j} x_{\text{ref}, i}, 
    \quad \text{if } |\mathcal{R}_j| > 0
    \label{eq:replacement}
\end{equation}

\paragraph{3) Adaptive Cluster Reinforcement (ACR):} Each $x_{\text{ref},i }$ is treated as an additional data point, and the $x_{\text{cluster},j}$ is updated via weighted averaging:
\begin{equation}
\begin{aligned}
    x_{\text{cluster}, j}^{\text{refined}} &= 
    \frac{|\mathcal{R}_j| \cdot \bar{x}_{\text{ref}, j} + |\mathcal{C}_j| x_{\text{cluster}, j}}{|\mathcal{R}_j| + |\mathcal{C}_j|}, \\
    \bar{x}_{\text{ref}, j} &= 
    \frac{1}{|\mathcal{R}_j|} \sum_{i \in \mathcal{R}_j} x_{\text{ref}, i}, 
    \quad \text{if } |\mathcal{R}_j| > 0
\end{aligned}
    \label{eq:merging}
\end{equation}
where $\mathcal{R}_j = \{ i \mid j^*_i = j \}$, and \( |\mathcal{C}_j| \) is the number of original tokens used to construct \( x_{\text{cluster}, j} \). 

In our ablation experiments in the \textit{Ablation Studies} section, we find that RBS (Eq.~\ref{eq:replacement}) is more effective at early clustering locations, while CGA (Eq.~\ref{eq:averaging}) performs best at later clustering locations. Therefore, we introduce a refinement method switching parameter $d$, where RBS is used as the refinement method when token clustering occurs before $d$ and CGA is used otherwise. Please refer to Algorithm 1 in supplementary material for the pseudocode. 

We also explore two additional strategies to improve the temporal element of our method. We first investigate a dynamic keyframe interval strategy to allow a more flexible keyframe selection scheme. Secondly, inspired by recent work in non-parametric memory consolidation~\cite{balavzevic2024memory}, we introduce a cluster memory module to enhance temporal correspondence. However, neither strategy yields noticeable performance improvements, therefore, we do not incorporate them into our final method.\footnote{See \textit{Discarded Ideas} in Supplementary for detailed analysis.}

\begin{table*}[t]
\centering
\caption{Main results on the YouTube-VIS 2019 and 2021 validation sets. Results improved by applying our TCA are shown in \underline{\textbf{bold}}. For a fair comparison, TCA settings are deliberately selected to maintain FPS equal to or higher than their counterparts.\protect\footnotemark}
\label{tab:ytvis19-21}
\resizebox{\textwidth}{!}{%
\begin{tabular}{c|l|ccccccc|ccccccc}
\toprule
\multirow{2}{*}{\textbf{Config}} 
& \multirow{2}{*}{\textbf{Method}} 
& \multicolumn{7}{c|}{\textbf{YTVIS19}} 
& \multicolumn{7}{c}{\textbf{YTVIS21}} \\
\cline{3-16}
& & AP $\uparrow$& AP$_{50}$ $\uparrow$& AP$_{75}$ $\uparrow$& AR$_{1}$ $\uparrow$& AR$_{10}$ $\uparrow$& FPS $\uparrow$& GFLOPs $\downarrow$
  & AP $\uparrow$& AP$_{50}$ $\uparrow$& AP$_{75}$ $\uparrow$& AR$_{1}$ $\uparrow$& AR$_{10}$ $\uparrow$& FPS $\uparrow$& GFLOPs $\downarrow$ \\
\midrule
\multicolumn{2}{l|}{\textit{Baseline (MinVIS)} \cite{huang2022minvis}} 
& 62.0 & 84.3 & 68.6 & 55.0 & 67.1 & 23.4 & 305.0 
& 55.8 & 77.3 & 63.0 & 46.8 & 60.9 & 23.7 & 302.1 \\
\midrule
\multirow{4}{*}{\textbf{Highest}} 
& MaskVD \cite{sarkar2024maskvd}
& 56.6 & 80.9 &63.6  &52.0 &61.3  &26.0  & 214.7
&  50.4& 74.8 & 56.7 & 43.8& 56.0 & 26.6 &  211.5\\ 
& ToMe \cite{bolya2022token}
& 58.8 & 81.3 & 64.8 & 51.7 & 63.7 & 26.0 & 272.8
& 53.8 & 75.5 & 60.3 & 45.0 & 59.2 & 26.2 & 270.3 \\
\cline{2-16}
& AiluRus \cite{li2023ailurus}
& 59.7 & 83.2 & 67.0 & 52.4 & 64.8 & 27.4 & 247.0
& 53.4 & 76.2 & 59.3 & 44.7 & 58.9 & 27.3 & 257.1 \\
& +Ours
& \underline{\textbf{61.0}} & \underline{\textbf{85.1}} & \underline{\textbf{68.0}} & \underline{\textbf{54.0}} & \underline{\textbf{66.4}} & 27.4 & 247.7
& \underline{\textbf{54.4}} & \underline{\textbf{76.4}} & \underline{\textbf{62.7}} & \underline{\textbf{45.5}} & \underline{\textbf{59.7}} & 27.3 & 257.7 \\
& Expedit \cite{liang2022expediting}
& 60.5 & 84.5 & 67.0 & 53.7 & 65.7 & 26.3 & 265.9
& 55.8 & 77.8 & 62.2 & 46.0 & 61.0 & 26.6 & 256.2 \\
& +Ours
& \underline{\textbf{61.7}} & \underline{\textbf{85.4}} & \underline{\textbf{69.1}} & \underline{\textbf{54.4}} & \underline{\textbf{67.1}} & 26.3 & 255.1
& \underline{\textbf{56.8}} & \underline{\textbf{78.6}} & \underline{\textbf{63.8}} & \underline{\textbf{47.0}} & \underline{\textbf{61.9}} & 26.6 & 257.7 \\
\midrule
\multirow{4}{*}{\shortstack{ \textbf{30\%} \\ \textbf{Speed Up} }} 
& MaskVD \cite{sarkar2024maskvd}
& 38.5 & 63.2 & 39.8 &36.8 &43.7  &29.5  &  159.0
&  34.8&  57.3& 37.1 & 32.3& 39.8 & 30.2 & 155.4 \\ 
& ToMe \cite{bolya2022token}
& 51.8 &77.8  &57.1  &47.3 &56.9  & 30.3 & 226.7 
& 43.2 & 66.9 & 45.7 & 37.5& 48.8 & 30.5 &  236.1\\ 

\cline{2-16}
& AiluRus \cite{li2023ailurus}
& 53.8 & 76.5 & 60.1 & 49.0 & 58.7 & 30.3 & 218.7
& 44.9 & 67.6 & 47.7 & 39.5 & 50.2 & 30.8 & 216.6 \\
& +Ours
& \underline{\textbf{56.4}} & \underline{\textbf{79.1}} & \underline{\textbf{63.4}} & \underline{\textbf{50.8}} & \underline{\textbf{61.6}} & 30.3 & 219.3
& \underline{\textbf{49.6 }}& \underline{\textbf{71.4}} & \underline{\textbf{55.8}} & \underline{\textbf{42.3}} & \underline{\textbf{55.0}} & 31.5 & 208.5 \\
& Expedit  \cite{liang2022expediting}
& 57.8 & 82.6 & 62.7 & 51.9 & 63.4 & 30.6 & 210.1
& 53.6 & 75.9 & 60.1 & 44.5 & 59.0 & 30.2 & 216.7 \\
& +Ours
& \underline{\textbf{59.3}} & \underline{\textbf{83.5}} & \underline{\textbf{67.0}} & \underline{\textbf{52.7}} & \underline{\textbf{64.3}} & 30.6 & 210.6
& \underline{\textbf{54.0}} & 75.6 & \underline{\textbf{61.7}} & \underline{\textbf{45.1}} & \underline{\textbf{59.5}} & 30.8 & 208.6 \\
\midrule
\multirow{4}{*}{\shortstack{ \textbf{50\%} \\ \textbf{Speed Up} }}
& MaskVD \cite{sarkar2024maskvd}
& 19.8 & 45.4 & 14.6 & 21.0&24.6  &34.4  &79.9 
& 24.0 & 51.5 &18.1  & 23.5&28.7 & 34.7 &  98.3\\ 
& ToMe \cite{bolya2022token}
& 37.6 &61.8  &40.1  &35.2 & 42.2 &34.9  &  190.3
& 34.0 & 56.6 & 37.3 & 31.5& 39.6 & 35.0 &  202.7\\ 
\cline{2-16}
& AiluRus \cite{li2023ailurus}
& 38.0 & 61.8 & 38.0 & 36.4 & 43.4 & 34.8 & 183.9
& 30.0 & 52.0 & 29.9 & 27.5 & 35.3 & 35.3 & 182.2 \\
& +Ours
& \underline{\textbf{49.9}} & \underline{\textbf{73.7}} & \underline{\textbf{54.7}} & \underline{\textbf{46.3}} & \underline{\textbf{55.3}} & 34.9 & 184.5
& \underline{\textbf{39.4}} & \underline{\textbf{61.8}} & \underline{\textbf{42.9}} & \underline{\textbf{35.0}} & \underline{\textbf{43.9}} & 36.8 & 167.4 \\
& Expedit  \cite{liang2022expediting}
& 42.5 & 66.8 & 45.5 & 39.4 & 47.8 & 35.2 & 175.3
& 36.8 & 58.9 & 39.0 & 31.7 & 41.6 & 35.9 & 173.7 \\
& +Ours
& \underline{\textbf{48.0}} & \underline{\textbf{72.4}} & \underline{\textbf{51.7}} & \underline{\textbf{45.3}} & \underline{\textbf{53.6}} & 35.6 & 166.7
& \underline{\textbf{40.2}} & \underline{\textbf{62.4}} & \underline{\textbf{43.6}} & \underline{\textbf{36.4}} & \underline{\textbf{45.3}} & 35.9 & 173.9 \\
\bottomrule
\end{tabular}%
}
\end{table*}

\begin{table}[ht!]
\centering
\caption{Main results on the OVIS validation sets. }
\label{tab:ovis}
\resizebox{0.48\textwidth}{!}{%
\begin{tabular}{c|l|ccccccc}
\toprule
\multirow{2}{*}{\textbf{Config}} 
& \multirow{2}{*}{\textbf{Method}} 
& \multicolumn{7}{c}{\textbf{OVIS}} \\
\cline{3-9}
& & AP $\uparrow$& AP$_{50}$ $\uparrow$& AP$_{75}$ $\uparrow$& AR$_{1}$ $\uparrow$& AR$_{10}$ $\uparrow$& FPS $\uparrow$& GFLOPs $\downarrow$\\
\midrule
\multicolumn{2}{l|}{\textit{Baseline (MinVIS)} } 
& 41.3 & 64.5 & 42.9 & 18.6 & 44.6 & 24.9 & 278.2 \\
\midrule
\multirow{4}{*}{\textbf{Highest}}
& MaskVD 
&  29.3&  50.5&30.0  &14.0 &22.0  &29.4  & 185.6\\ 
& ToMe
& 38.6 & 62.7 & 39.4 & 18.0 & 41.9 & 28.4 & 248.8 \\
\cline{2-9}
& AiluRus  
& 38.1 & 62.5 & 38.1 & 17.9 & 41.3 & 29.5 & 236.6 \\
& +Ours
& \underline{\textbf{38.9}} & 61.8 & \underline{\textbf{39.6}} 
& \underline{\textbf{18.1}} & \underline{\textbf{42.4}} 
& 29.6 & 237.7 \\
& Expedit 
& 38.9 & 61.9 & 39.3 & 17.3 & 42.5 & 29.6 & 231.4 \\
& +Ours
& \underline{\textbf{39.5}} & 61.4 & \underline{\textbf{41.6}} 
& \underline{\textbf{17.8}} & \underline{\textbf{42.7}} 
& 30.4 & 216.2 \\
\midrule
\multirow{4}{*}{\shortstack{ \textbf{30\%} \\ \textbf{Speed Up} }}
& MaskVD 
&  20.2&39.9  &19.0 & 8.9& 12.4& 32.1   & 145.1\\ 
& ToMe
&  31.5&  57.0& 31.7 &14.2 &24.5  & 31.8  & 239.8\\ 
\cline{2-9}
& AiluRus 
& 31.5 & 53.0 & 31.2 & 15.8 & 35.2 & 32.1 & 207.3 \\
& +Ours
& \underline{\textbf{32.5}} & \underline{\textbf{56.1}} 
& \underline{\textbf{31.8}} & 15.8 
& \underline{\textbf{36.2}} & 32.4 
& 206.3 \\
& Expedit 
& 36.0 & 60.1 & 36.9 & 16.7 & 39.2 & 32.1 & 199.6 \\
& +Ours
& \underline{\textbf{38.7}}& \underline{\textbf{63.8}} 
& \underline{\textbf{39.9}} & \underline{\textbf{18.0}} 
& \underline{\textbf{42.0}} & 32.3
& 200.5 \\
\midrule
\multirow{4}{*}{\shortstack{ \textbf{50\%} \\ \textbf{Speed Up} }}
& MaskVD 
& 5.7 & 14.3 & 3.7 &2.0 &3.7  &  36.8& 85.0\\ 
& ToMe 
& 12.5 & 27.5 & 10.5 &5.3 & 7.3 & 36.1 &193.6 \\ 
\cline{2-9}
& AiluRus 
& 15.6 & 33.6 & 13.1 & 8.9 & 19.8 & 36.8 & 175.6 \\
& +Ours
& \underline{\textbf{22.5}} & \underline{\textbf{43.4}} 
& \underline{\textbf{21.0}} & \underline{\textbf{12.5}} 
& \underline{\textbf{26.3}} & 38.1
& 167.8 \\
& Expedit 
& 16.4 & 33.3 & 13.4 & 9.5 & 19.5 & 38.6 & 159.9 \\
& +Ours
& \underline{\textbf{20.8} }& \underline{\textbf{40.3} }
& \underline{\textbf{19.6} }& \underline{\textbf{12.3} }
& \underline{\textbf{23.7} }& 39.0 & 150.8 \\
\bottomrule
\end{tabular}%
}

\end{table}

\section{Experiments}

\subsection{Datasets \& Implementation Details}
We evaluate our method on the YouTube-VIS (YTVIS) 2019, YouTube-VIS 2021 \cite{yang2019video}, and Occluded VIS (OVIS) \cite{qi2022occluded} datasets. YTVIS 2019 and 2021 contain 40 object categories, with YTVIS 2021 offering higher-quality annotations. OVIS is a more challenging dataset with 25 categories, focusing on frequent partial or full occlusions.  It also features longer sequences than YTVIS, making it a more demanding benchmark.

To further evaluate the applicability of TCA, we conduct a VSS case study on a private surgical dataset from laparoscopic cholecystectomy procedures. The dataset contains 1966 videos of 30-second sequences, annotated at 1 FPS, with pixel-level masks for the cystic artery and cystic duct; other structures are labeled as background. The dataset is split into 1641 training, 104 validation, and 221 test videos.

We use the  official MinVIS \cite{huang2022minvis} implementation with a pretrained Swin Large backbone. All hyperparameters are kept at their default. TCA is applied to both token clustering methods: Expedit~\cite{liang2022expediting} and AiluRus~\cite{li2023ailurus}. Inputs are resized to a minimum size of 480 pixels. Following standard VIS validation practice \cite{huang2022minvis, heo2022vita,heo2023generalized}, all results are reported on the validation set.

For the surgical case study, we use a Swin  Large backbone followed by a three-layer convolutional decoder (SwinSeg). The model is trained on the training set with early stopping based on validation set performance. Results are reported on the test set. All images are resized to 576 × 320.

Across all datasets, inputs are processed in an online setting with a batch size of 1. All experiments are conducted on a single A6000 GPU. We empirically set the reference offset $\beta$ and keyframe interval $f_{\text{max}}$ to 6, and the refinement switching parameter $d$ to 2. Experiments sweep all combinations of cluster size $N$ from $8 \times 8$ to $2 \times 2$, and  clustering location $\alpha$ from 0 to 10 within Stage 3 of Swin. Comparisons are reported only for Pareto-optimal results. Unless specified otherwise, ablations use YTVIS19 with Expedit and  $N=5 \times 5$. 

Following prior VIS works \cite{huang2022minvis, heo2022vita, heo2023generalized}, we report Average Precision (AP) and Average Recall (AR) on VIS datasets. AP measures segmentation quality as the area under the precision–recall curve across IoU thresholds, while AR evaluates instance retrieval at a fixed number of instances per video. IoU sums intersections over unions across frames to capture both spatial and temporal accuracy. For the surgical dataset, we use mean IoU (mIoU) for evaluation. Efficiency is measured by throughput in FPS and FLOPs, averaged over the test set.

\subsection{Main results}

We evaluate the proposed TCA by integrating it into the two SOTA token clustering methods---Expedit and AiluRus. Table~\ref{tab:ytvis19-21} presents results on YTVIS 2019 and YTVIS 2021 across three settings: (1) the highest achievable AP with token clustering, (2) a 30\% FPS increase over the baseline, and (3) a 50\% FPS increase. For each method, we select configurations along the Pareto frontier that most closely match the target speedup levels. When evaluating TCA, we similarly identify configurations that achieve equal or higher FPS compared to their non-TCA counterparts. To further compare TCA against SOTA compression methods, we also include ToMe, a widely used token merging method, and MaskVD, the current SOTA for efficient online video detection. All methods are applied to MinVIS.

\begin{figure*}[t]
\begin{center}
   \includegraphics[width=\linewidth]{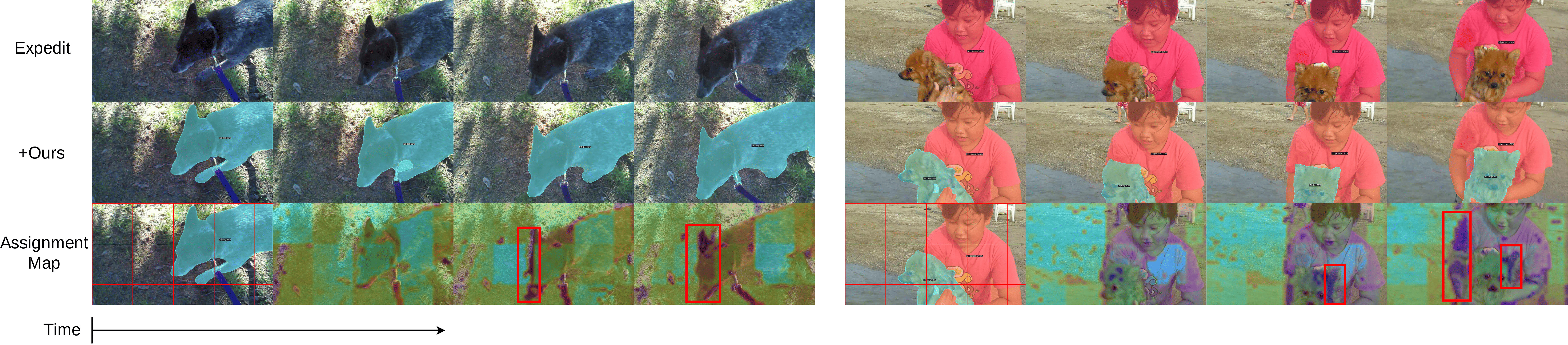}
\end{center}
   \caption{Qualitative comparisons of predictions on YTVIS 2019 generated using: (Top) Expedit only, (Middle) Expedit with the addition of our TCA, and (Bottom) window partitions for reference frame and the corresponding assignment heatmap for non-reference frame. Darker colors indicates a lower frequency of assignment. TCA enables refinement of the majority of tokens that share similar features with the reference frame window (light green area), while recognizing objects that are not present in the reference window (dark areas highlighted in red boxes). }
\label{comparison_vis}
\end{figure*}

\begin{figure*}[t]
\begin{center}
   \includegraphics[width=\linewidth]{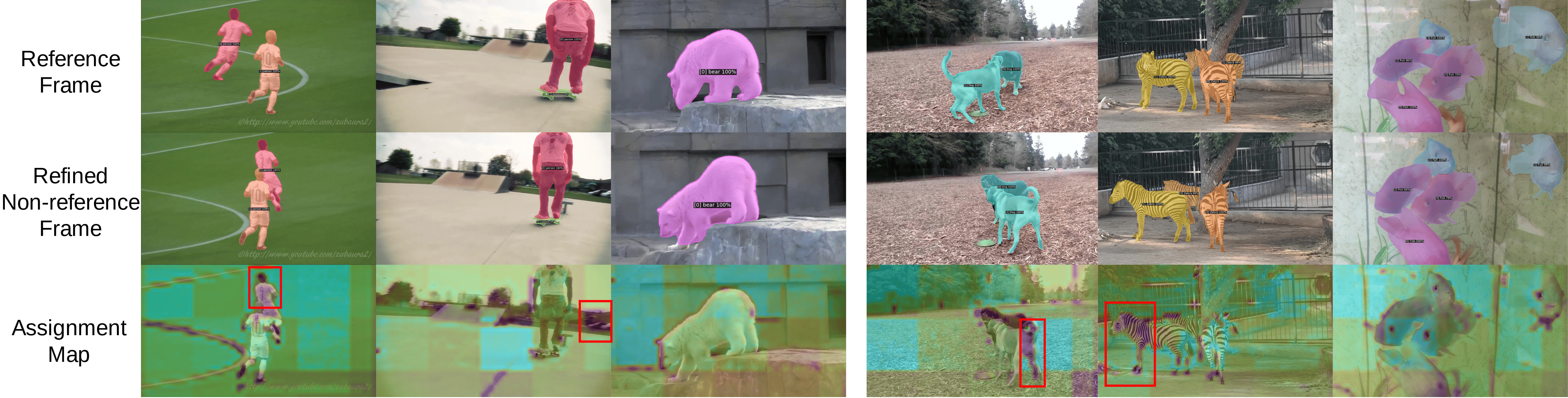}
\end{center}
   \caption{Visualization of the assignment heatmap on (Left) YouTube-VIS 2021 and (Right) OVIS.}
\label{assignment_vis}
\end{figure*}

\footnotetext{See Supplementary Tab. A1 for $N$ and $\alpha$ used in each setting.}

In all settings, TCA consistently improves the performance of Expedit and AiluRus across all metrics when measured at an equivalent level of speedup. Improvements are most pronounced under 50\% speedup, where TCA yields an average AP improvement of 10.7 for AiluRus and 4.5 for Expedit, and an average AR$_{1}$ improvement of 8.7 for AiluRus and 5.3 for Expedit.  In YTVIS21, TCA also outperform MinVIS by 1.0 AP while offering 12\% faster FPS and 15\% fewer FLOPs. Compared to ToMe, both Expedit and AiluRus consistently deliver stronger performance—especially when combined with TCA.  In contrast, MaskVD relies on expensive scatter/gather operations to maintain window-attention compatibility. These memory-bound operations, although not captured in FLOPs, lead to substantial latency overhead \cite{dutson2023eventful, sarkar2024maskvd}. To compensate, MaskVD requires more aggressive token compression, artificially lowering GFLOPs without achieving tangible efficiency gains. The observed poor performance suggests this design does not translate well to video segmentation. In comparison, integrating TCA into Expedit or AiluRus significantly outperforms MaskVD, demonstrating that TCA is a more effective approach than directly adapting  detection-oriented methods to video segmentation.

Similar trends are observed on the OVIS benchmark (Table~\ref{tab:ovis}), a more challenging dataset due to its complex object motions and occlusions. TCA consistently enhances the performance of both token clustering methods across all settings. At the 50\% speedup level, TCA yields an AP improvement of 6.9 for AiluRus and 4.4 for Expedit, further demonstrating the robustness and generalizability of our approach.

Whilst our main experiments focus on Swin-based backbones, we believe the underlying design of TCA is applicable to other tasks and Transformer architectures such as ViT. As an initial step, we present a preliminary result on applying TCA to a video dense prediction task using a ViT backbone in the supplementary material, and we leave a more systematic investigation to future work.

\subsection{Qualitative Results}

We highlight the benefits of TCA by visualizing the VIS prediction masks produced by Expedit alone and with TCA. Fig.~\ref{comparison_vis} shows a prediction comparison at $\alpha = 0$ on YTVIS 2019. In the bottom row, for the reference frame (first frame), we illustrate how the image is partitioned into windows in stage 3 of Swin and display the assignment heatmap in subsequent non-key frames, which indicates the frequency with which each patch is matched to a reference token. In both scenes, an overall uniform rate of assignment is observed in regions with minimal changes, such as the background. Across frames, certain object parts—such as the dog’s body in both clips and the person's shoulder in the right clip (see red boxes)---move across window boundaries relative to the reference frame. In the assignment map, these regions appear as dark spots, indicating a low assignment rate. This suggests that the assignment operates correctly, as no similar tokens exist within the reference window, which primarily contains background tokens.

Fig.~\ref{assignment_vis} presents additional assignment heatmaps in YTVIS 2021 and OVIS. Similar to Fig. \ref{comparison_vis}, objects absent in the reference window—such as the top person in the first image, the cars in the second image, the right foot of the dog in fourth image, and the top half of the left zebra in the fifth image, exhibit low assignment rates (see red box). Conversely, objects that remain within the same window (the bottom person in the first image, the lower half of the left zebra in the fifth image) and frames with minimal motion (the third and last image), display higher assignment rates, indicated by brighter colors. These patterns confirm that TCA is capable of refining similar cluster tokens to tokens within reference windows, while leaving dissimilar clusters unmodified.


\subsection{Ablation Studies}
\label{ablation}

\begin{figure*}[t]
\centering
    \begin{subfigure}[b]{0.32\linewidth}
        \centering
        \includegraphics[width=\linewidth]{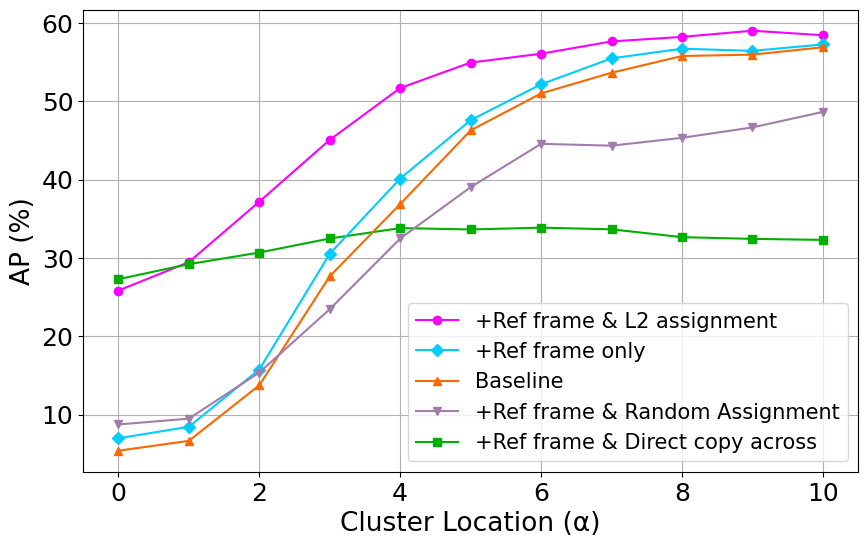}
        \caption{}
        \label{fig:assignment_ablation1}
    \end{subfigure}
    \hfill
    \begin{subfigure}[b]{0.32\linewidth}
        \centering
        \includegraphics[width=\linewidth]{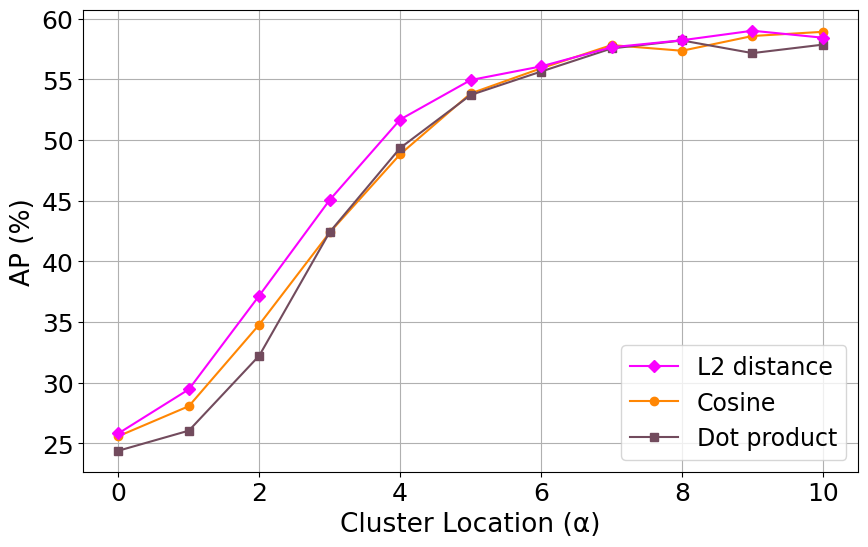}
        \caption{}
        \label{fig:assignment_ablation2}
    \end{subfigure}
    \hfill
    \begin{subfigure}[b]{0.32\linewidth}
        \centering
        \includegraphics[width=\linewidth]{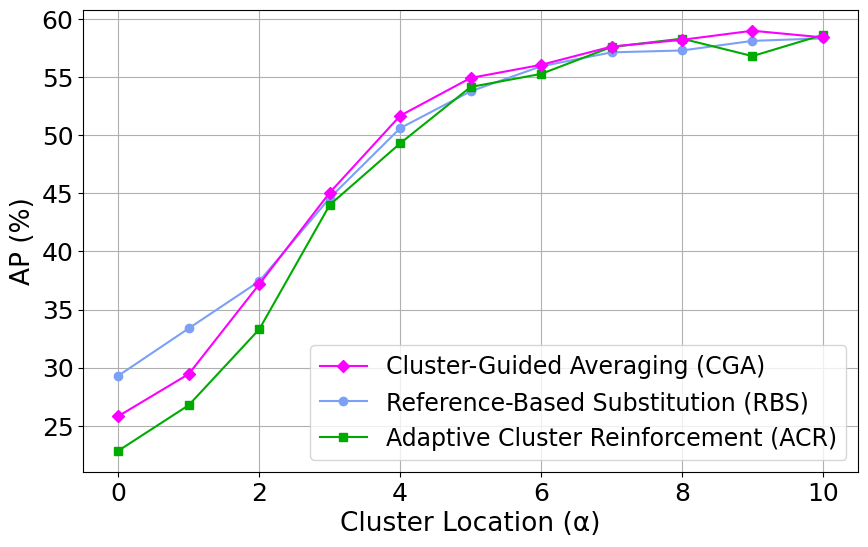}
        \caption{}
        \label{fig:assignment_ablation3}
    \end{subfigure}
    
    \caption{Ablation results for: (a) Improvement contribution, (b) Similarity metric for cluster assignment, (c) Refinement method for cluster refinement.}
    \label{fig:assignment_ablation}
\end{figure*}


Due to the temporal nature of the AP metric, the mere addition of a reference frame, which performs late clustering, can provide some performance improvement. To confirm that the majority of the performance gain comes from the assignment and refinement of clusters in the TCA layer, rather than just the presence of a reference frame, Fig.~\ref{fig:assignment_ablation1} presents an ablation study across different clustering locations $\alpha$. The results show that adding a reference frame without performing assignment and refinement yields only a slight improvement over using Expedit alone. Additionally, both directly copying reference features across frames and random assignment negatively impact performance. In contrast, L2-based assignment yields the most significant improvement, demonstrating that the TCA layer  is the primary contributor to the observed performance gain.


Fig.~\ref{fig:assignment_ablation2} presents an ablation study comparing three similarity metrics used to compute $j^*$ between reference and cluster tokens. L2 distance demonstrates a consistent advantage across all $\alpha$. Given that all three metrics have similar computational costs, we ultimately choose L2 distance.



Fig.~\ref{fig:assignment_ablation3} presents an ablation study on the refinement method applied to clustered tokens. RBS demonstrates a clear advantage at early clustering locations but deteriorates as $\alpha$ increases. We hypothesize that this is due to the lower reliability of token features in early blocks, which leads to noisy clusters. As a result, directly incorporating these clusters negatively impacts performance. As the clustering location moves deeper, cluster reliability improves, and incorporating these clusters provides valuable information not present in the reference tokens, leading to better performance. Based on this observation, we use RBS when the clustering location is smaller than 2 and CGA otherwise ($d = \text{2}$).

\begin{table}[t]
    \centering
    \resizebox{0.48\textwidth}{!}{
    \begin{tabular}{c c c c c c c c}
        \toprule
        \textbf{Reference Offset $\beta$} & \textbf{AP}$\uparrow$ & \textbf{AP$_{50}$}$\uparrow$ & \textbf{AP$_{75}$}$\uparrow$ & \textbf{AR$_1$}$\uparrow$ & \textbf{AR$_{10}$}$\uparrow$ &  \textbf{FPS}$\uparrow$ & \textbf{GFLOPs}$\downarrow$ \\
        \midrule
        $3$ & 57.0 & 81.4& 63.1& 51.3& 62.4& 30.8 & 210.4\\
        $4$ & 57.6 & 81.6& 64.1& 51.7& 62.8& 30.6& 211.9\\
        $5$ & 58.2 & 82.2& 65.0& 52.0& 63.6& 30.4 & 213.4\\
        
        $6$ & 58.3 & 82.5& 65.0& 52.1& 63.6& 30.3 & 215.0\\
        $7$ & 58.4 & 82.5& 65.1& 52.3& 63.7& 30.1 & 216.5\\
        \bottomrule
    \end{tabular}
    }
    \caption{Effect of reference offset $\beta$ on performance}
    \label{tab:ref_offset}
\end{table}

\begin{table}[t]
    \centering
    \resizebox{0.48\textwidth}{!}{
    \begin{tabular}{c c c c c c c c}
        \toprule
        \textbf{Keyframe Interval $f_{\text{max}}$} & \textbf{AP}$\uparrow$ & \textbf{AP$_{50}$}$\uparrow$ & \textbf{AP$_{75}$}$\uparrow$ & \textbf{AR$_1$}$\uparrow$ & \textbf{AR$_{10}$}$\uparrow$ &  \textbf{FPS}$\uparrow$ & \textbf{GFLOPs}$\downarrow$ \\
        \midrule
        $3$ & 58.3 & 82.4& 65.3& 52.5&  63.9& 29.4 & 223.5\\
        $4$ & 58.1 & 81.1& 64.9& 52.1&  63.5& 29.9 & 219.1\\
        $5$ & 58.1 & 81.9& 65.0& 52.0&  63.5& 30.1 & 216.9\\
        
        $6$ & 58.3 & 82.5& 65.0& 52.1&  63.6 & 30.3 & 215.0\\
        $7$ & 57.4 & 80.5& 64.2 &51.8 &  63.0& 30.4 & 214.3\\
        \bottomrule
    \end{tabular}
    }
    \caption{Effect of keyframe interval $f_{\text{max}}$ on performance}
    \label{tab:reset_frame}
\end{table}

We also perform ablation studies on the reference offset $\beta$ and keyframe interval $f_{\textbf{max}}$, using a clustered token size $N$ of $6 \times 6$ and report the average performance across all $\alpha$.
Table~\ref{tab:ref_offset} presents the effect of varying $\beta$. As $\beta$ increases, reference tokens remain fully processed for a longer duration, leading to improved quality at the cost of higher computation. However, there is a diminishing return as $\beta$ increases, therefore we select $\beta=6$ to balance this trade-off. Table~\ref{tab:reset_frame} evaluates the impact of varying $f_{\text{max}}$. As $f_{\text{max}}$ increases, FPS improves due to the reduced processing of reference frames. However, features may drift as objects move within the scene and background changes occur, making reference tokens less reliable over time. Performance remains stable up to an interval of $7$, after which a significant $0.9$ AP drop is observed. To achieve the best trade-off, we set $f_{\text{max}}=6$.

\subsection{Case Study: Surgical Dataset}
\begin{figure}[t]
\begin{center}
   \includegraphics[width=\linewidth]{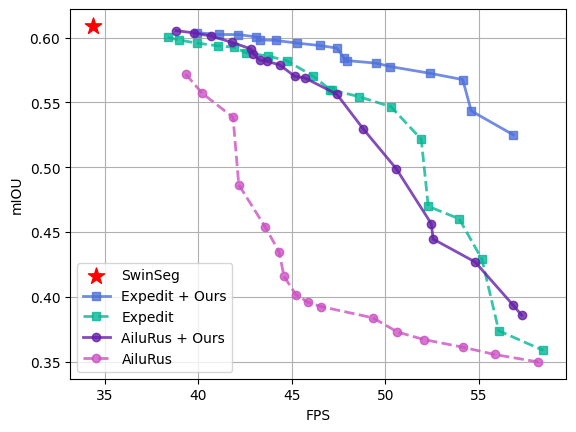}
\end{center}
   \caption{Pareto frontier illustrating the optimal mIoU–FPS trade-off on the surgical dataset. Incorporating our method consistently improves performance across all settings.}
\label{jura_pareto}
\end{figure}

To assess the applicability of TCA beyond natural videos, we explore the task of VSS using a private surgical dataset. Fig.~\ref{jura_pareto} illustrates the Pareto curve of performance; a similar trend to YTVIS19 (Fig.~\ref{ytvis19_pareto}) is observed. In both Expedit and AiluRus, the addition of TCA leads to significant improvements. Specifically, using only Expedit, at high compression with $N = 5 \times 5$ and $\alpha = 0$ (second-to-last point), yields a $63\%$ FPS improvement but results in a noticeable drop in mIoU from 0.609 to 0.374. In contrast, with TCA, we achieve a similar speedup ($66\%$, last point in curve) while maintaining a significantly higher mIoU of 0.525 (+0.151). For visualizations, see Fig.~C in the supplementary.


\section{Conclusion}
We introduce Temporal Cluster Assignment (TCA), a fine-tuning-free method that enhances token clustering for video segmentation by leveraging temporal correlations. TCA refines token clusters across frames, enabling a more favorable accuracy–speed trade-off without additional training. Extensive experiments on YouTube-VIS 2019/2021, OVIS, and a private surgical vision dataset show that integrating TCA into existing token clustering methods consistently improves their efficiency-accuracy trade-off across diverse settings. These results highlight the potential of TCA for deployment in real-time, resource-constrained video segmentation applications across both natural and specialized domains.

\newpage
\bibliography{aaai2026}

\newpage

\maketitle

\subsection{Preliminary Extension to ViT-Based Depth Estimation}

\begin{table}[H]
    \centering
    \resizebox{0.48\textwidth}{!}{
    \begin{tabular}{ c c c c c c c c}
        \toprule
        \textbf{Method}
          & \textbf{RMSE}$\downarrow$
          & \textbf{RMSE$_{\log}$}$\downarrow$
          & \textbf{AbsRel}$\downarrow$
          & \textbf{SqRel}$\downarrow$
          & \textbf{SILog}$\downarrow$
          & \textbf{$\log_{10}$}$\downarrow$ 
          & \textbf{FPS}$\uparrow$\\
        \midrule
        Baseline (DPT) &2.573 &0.0922&0.0624  & 0.222 & 8.284 & 0.0271  & 30.59 \\
        \hline
        Expedit &2.745 &0.0962&0.0638  & 0.242 & 8.687 & 0.0278  & 35.63 \\
        +Ours & \underline{\textbf{2.682}}  & \underline{\textbf{0.0948}}  &\underline{\textbf{0.0631} } & \underline{\textbf{0.235}} & \underline{\textbf{8.545}} & \underline{\textbf{0.0275}} & 36.34\\
        \bottomrule
    \end{tabular}
    }
    \caption{Preliminary depth estimation results using a ViT-style (DPT) model with and without TCA.}
    \label{tab:depth_metrics}
\end{table}
To explore whether TCA can generalize beyond window-based segmentation transformers like Swin, we conduct a preliminary experiment on a video depth estimation task with the KITTI dataset \cite{geiger2012we} using DPT \cite{ranftl2021vision}, which uses a ViT-style backbone with global attention.

As shown in Table~\ref{tab:depth_metrics}, integrating TCA into Expedit leads to an improvement in performance across standard depth metrics, alongside a slight increase in FPS. This result is not meant to establish comprehensive generalization, but it provides early evidence that the core idea of TCA—leveraging temporal consistency to refine token usage—may also be applicable to other dense prediction tasks involving ViT-like models. A more systematic investigation is left to future work.

\newpage
\FloatBarrier
\section{Pseudo Code}
\begin{algorithm}[H]
\caption{Temporal Cluster Assignment}
\vspace{-5pt} 
\begin{tcolorbox}[
    colback=lightgray, arc=0mm, boxrule=0pt, width=\linewidth, 
    left=1pt, right=1pt, top=0pt, bottom=0pt] 
\begin{lstlisting}[mathescape=true]
def forward(x, $x_{\text{ref}}$, $\alpha$, $\beta$, frame_idx, $f_{\text{max}}$, d):
    '''
    x: current frame clustered tokens
    x_ref: reference frame tokens
    alpha: clustering location
    beta: delayed clustering location
    frame_idx: current frame index
    f_max: keyframe interval
    d: refinement switching parameter
    '''
    key_frame = (frame_idx % $f_{\text{max}}$ == 0)
    if key_frame:
        for i, block in enumerate(blocks):
            # Late clustering
            if i == $\alpha$ + $\beta$:
                # Update reference tokens
                x$_{ref}$ = x           
                x = clustering(x)   
            x = block(x)
    else:
        for i, block in enumerate(blocks):
            # Early clustering
            if i == $\alpha$:
                x = clustering(x)    
            elif i == $\alpha$ + $\beta$:
                # Temporal Cluster Assignment 
                # Eq. 1
                distance = distance_fn(x, x$_{ref}$)
                j = argmin(distance)
                if $\alpha$ <= d:
                    # Eq. 3
                    x = RBS(x, x$_{ref}$, j)
                else:
                    # Eq. 2
                    x = CGA(x, x$_{ref}$, j)
            x = block(x)
    x = reconstruct(x)
    frame_idx += 1
    return x, x$_{ref}$, frame_idx
\end{lstlisting}
\end{tcolorbox}
\vspace{-10pt}
\label{pesudo_code}
\end{algorithm}

\newpage
\FloatBarrier
\section{Discarded Ideas}
\subsection{Dynamic Keyframe Interval}
\begin{figure}[!th]
\begin{center}
   \includegraphics[width=\linewidth]{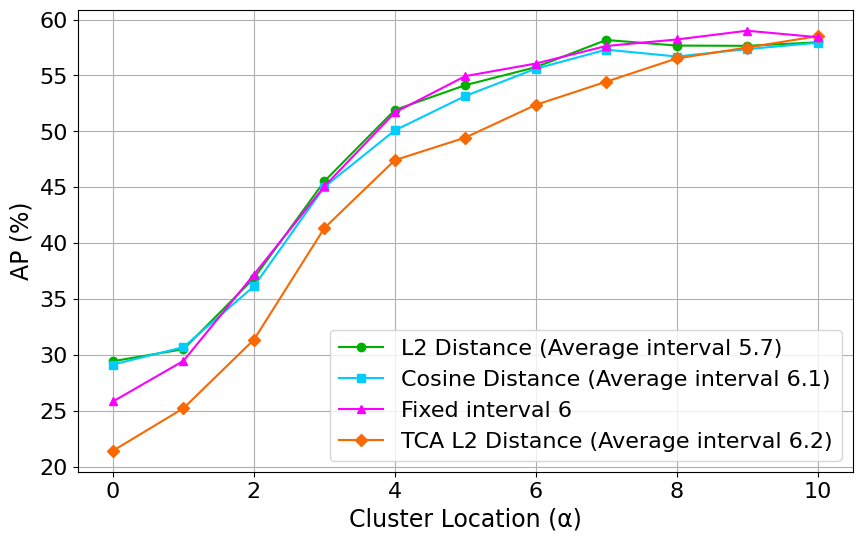}
\end{center}
   \caption{Comparison of dynamic keyframe interval schematic against fixed keyframe interval.}
\label{reset}
\end{figure}

Past work in feature propagation \cite{li2018low, zhu2018towards, xu2018dynamic} has proposed dynamic keyframe intervals based on a learned module. Given the similarity between our TCA and feature propagation, we also explored the possibility of a training-free dynamic keyframe interval based on existing properties of feature tokens. Specifically, we compute a similarity between current non-keyframe to the reference keyframe. This similarity score accumulates at each non-keyframe, and if the accumulated distance exceeds a predefined threshold, the next frame is considered as a keyframe. Specifically, we tested three different similarity functions:  

1) The average distance during the TCA assignment stage, as defined in the main text Eq. 1.

2) The cosine distance between reference frame tokens and current frame tokens, computed at the end of Stage 3 in the Swin Transformer.  

3) The L2 distance between reference frame tokens and current frame tokens, also computed at the end of Stage 3 in the Swin Transformer.  

Fig. \ref{reset} presents a comparison of all three dynamic interval methods against a fixed keyframe interval. While both cosine distance and L2 distance-based dynamic intervals show a slight advantage at $\alpha = 0$ and 1, none of the methods yield a noticeable improvement at later $\alpha$ values. Given the computational overhead required for distance calculations, we opted for a fixed keyframe interval instead.

\newpage
\subsection{Memory Module}
\label{sec:memory}
\begin{figure}[!th]
\begin{center}
   \includegraphics[width=\linewidth]{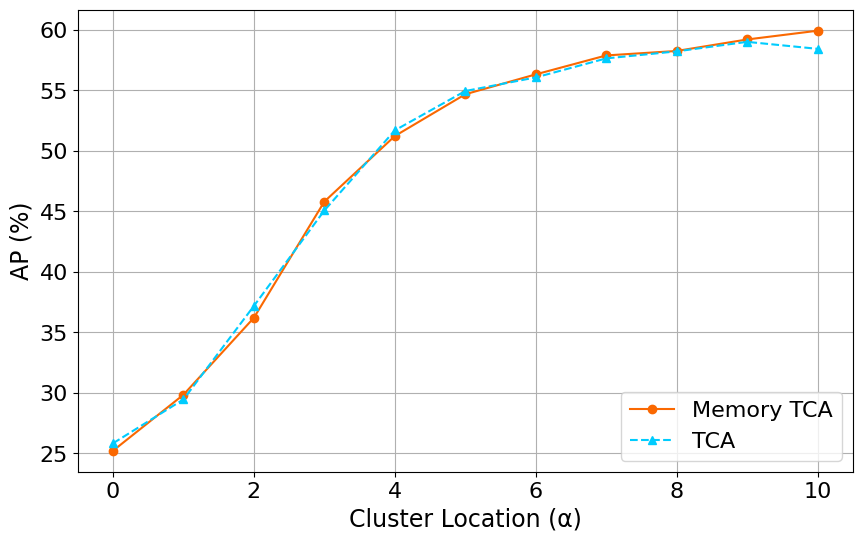}
\end{center}
   \caption{Effect of adding a memory module to TCA compared to using TCA alone.}
\label{memory}
\end{figure}

Inspired by recent methods in memory consolidation for long-context video understanding \cite{balavzevic2024memory}, we explored the addition of a non-parametric memory module during TCA. Specifically, we initialize a set of memory tokens $x_{\text{mem}}$ using the refined cluster $x_{\text{cluster}}^{\text{refined}} \in \mathbb{R}^{K \times N \times L}$ at the first non-keyframe, where $K$ is the number of windows, $N$ is the number of clustered tokens, and $L$ is the token dimension.

During TCA in subsequent frames, and before refinement, this memory is concatenated with the reference tokens along the window dimension, forming $x_{\text{ref}}^{\text{mem}} \in \mathbb{R}^{K \times (M+N) \times L}$, where $M$ represents the number of reference tokens. The concatenated tokens are then treated as the new reference tokens and used for assignment and refinement, following the original TCA process. Similar to $x_{\text{ref}}$, $x_{\text{mem}}$ are assigned and used to refine $x_{\text{cluster}}$. By taking into account features from past frames stored within,  $x_{\text{mem}}$ acts as a correction to feature drifting caused by differences between the reference and current frames, mitigating the impact from scene changes in video. 

In \cite{balavzevic2024memory}, the memory is updated by concatenating a k-means clustered form of its current token to the existing memory. Since both the k-means operation and the increase in memory size across frames cause extra overhead in TCA, we instead update the memory by directly replacing it  with the newly refined tokens $x_{\text{cluster}}^{\text{refined}}$ after the refinement step. This serves as a variant to \cite{balavzevic2024memory} where the previous memory is included into the newly updated memory through the assignment and refinement process.

Fig. \ref{memory} illustrates the effect of incorporating the memory module. While a slight advantage is observed at $\alpha = 10$, the memory module does not yield noticeable improvements at other values of $\alpha$. Given the additional computational overhead introduced by the increased concatenated reference size, we opt to use only the reference tokens in TCA.

\FloatBarrier
\begin{table*}[h!]
\centering
\caption{Setting of cluster size $N$ and cluster location $\alpha$ used in main results of Table 1 and 2.}
\label{tab:setting}
\resizebox{\textwidth}{!}{%
\begin{tabular}{c|l|cc|cc|cc}
\toprule
\multirow{2}{*}{\textbf{Config}} 
& \multirow{2}{*}{\textbf{Method}} 
& \multicolumn{2}{c|}{\textbf{YTVIS19}} 
& \multicolumn{2}{c|}{\textbf{YTVIS21}} 
& \multicolumn{2}{c}{\textbf{OVIS}} \\
\cline{3-8}
& & Cluster Size ($N$) & $\alpha$ & Cluster Size ($N$) & $\alpha$ & Cluster Size ($N$) & $\alpha$ \\
\midrule
\multicolumn{2}{l|}{\textit{Baseline (Unclustered size $M$)}} & 12 $\times$ 12 & / &12 $\times$ 12 &/ & 12 $\times$ 12 & / \\
\midrule
\multirow{4}{*}{\textbf{Highest}} 
& AiluRus & 7 $\times$ 7 & 10 &  8 $\times$ 8& 10&8 $\times$ 8  &10 \\
& +Ours & 7 $\times$ 7& 9&8 $\times$ 8 &9 &8 $\times$ 8  & 9\\
& Expedit & 8 $\times$ 8 & 10 & 8 $\times$ 8 & 10 & 8 $\times$ 8& 9 \\
& +Ours & 7 $\times$ 7 & 10 &  8 $\times$ 8  & 9 &6 $\times$ 6 &9 \\
\midrule
\multirow{4}{*}{\shortstack{ \textbf{30\%} \\ \textbf{Speed Up} }}
& AiluRus &8 $\times$ 8 & 7 &  6 $\times$ 6  & 8 & 6 $\times$ 6 &9 \\
& +Ours &  6 $\times$ 6 &7&  6 $\times$ 6 & 6 & 7 $\times$ 7 &6 \\
& Expedit & 6 $\times$ 6 & 7 & 6 $\times$ 6 & 8 &6 $\times$ 6 & 8\\
& +Ours &  6 $\times$ 6 & 6 &  6 $\times$ 6  & 6 & 6 $\times$ 6&7 \\
\midrule
\multirow{4}{*}{\shortstack{ \textbf{50\%} \\ \textbf{Speed Up} }} 
& AiluRus &  6 $\times$ 6 & 4& 6 $\times$ 6 & 4 & 6 $\times$ 6 & 5\\
& +Ours &  6 $\times$ 6  & 0&  5 $\times$ 5 & 3 & 6 $\times$ 6& 3\\
& Expedit &  6 $\times$ 6  & 3 &6 $\times$ 6 & 3 &6 $\times$ 6 & 3\\
& +Ours & 4 $\times$ 4 & 4 & 4 $\times$ 4 & 4 &4 $\times$ 4 & 4\\
\bottomrule
\end{tabular}%
}
\end{table*}

\begin{figure*}[!th]
\begin{center}
   \includegraphics[width=\linewidth]{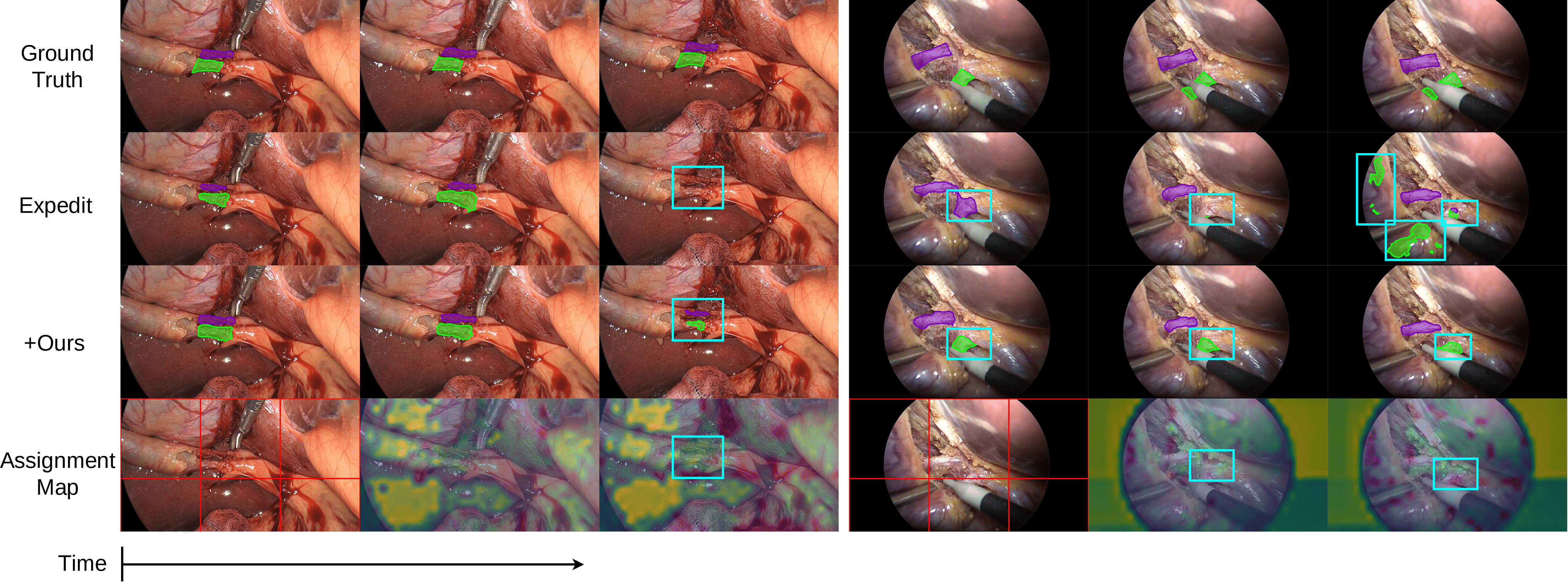}
\end{center}
   \caption{Qualitative comparisons of predictions on the surgical dataset.  Purple represents cystic duct and green represents cystic artery.  In both examples, using only Expedit (second row) results in incomplete or incorrect predictions, as indicated by the blue box. With the addition of TCA (third row), reference tokens refine the clustered tokens, correcting erroneous predictions. As seen in the bottom row, locations around incorrect predictions appear bright, indicating the evident contribution of the refinement to achieve the correct prediction.}
\label{jura_vis}
\end{figure*}

\begin{figure*}[t]
\begin{center}
   \includegraphics[width=\linewidth]{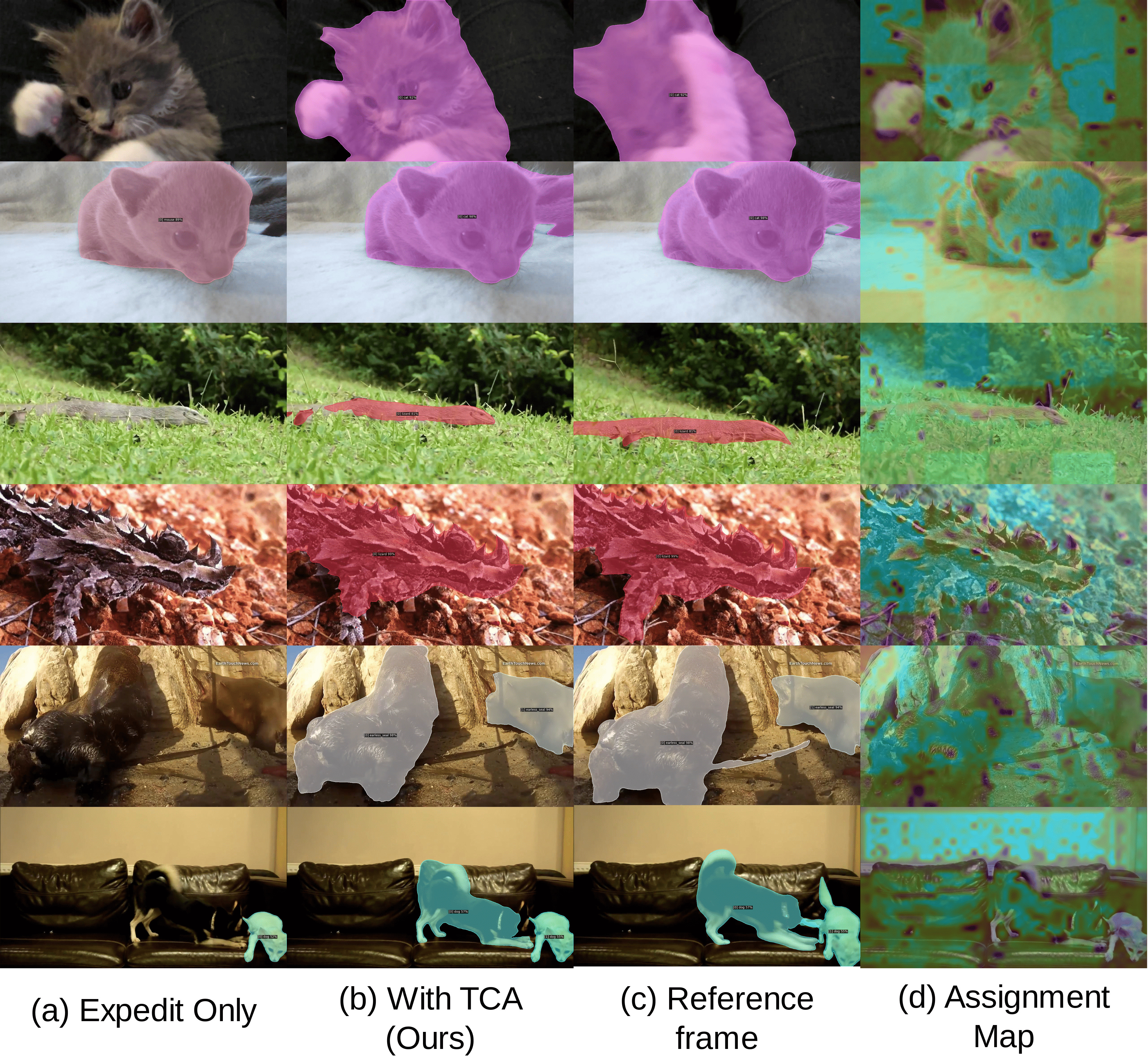}
\end{center}
   \caption{Further examples of comparisons and heatmaps from YTVIS19}
\label{fig:extra_19}
\end{figure*}

\begin{figure*}[t]
\begin{center}
   \includegraphics[width=\linewidth]{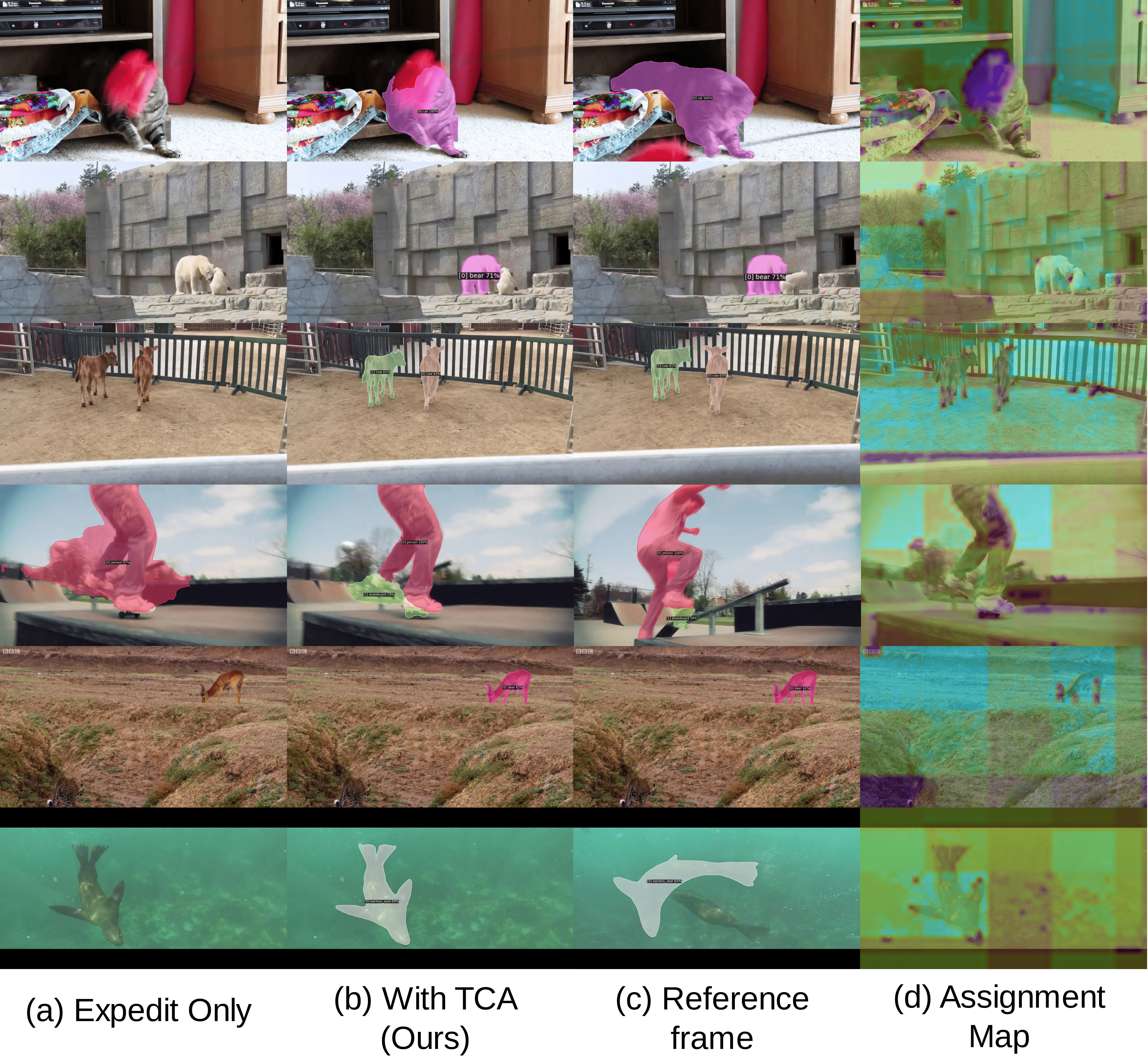}
\end{center}
   \caption{Further examples of comparisons and heatmaps from YTVIS21}
\label{fig:extra_21}
\end{figure*}

\begin{figure*}[t]
\begin{center}
   \includegraphics[width=\linewidth]{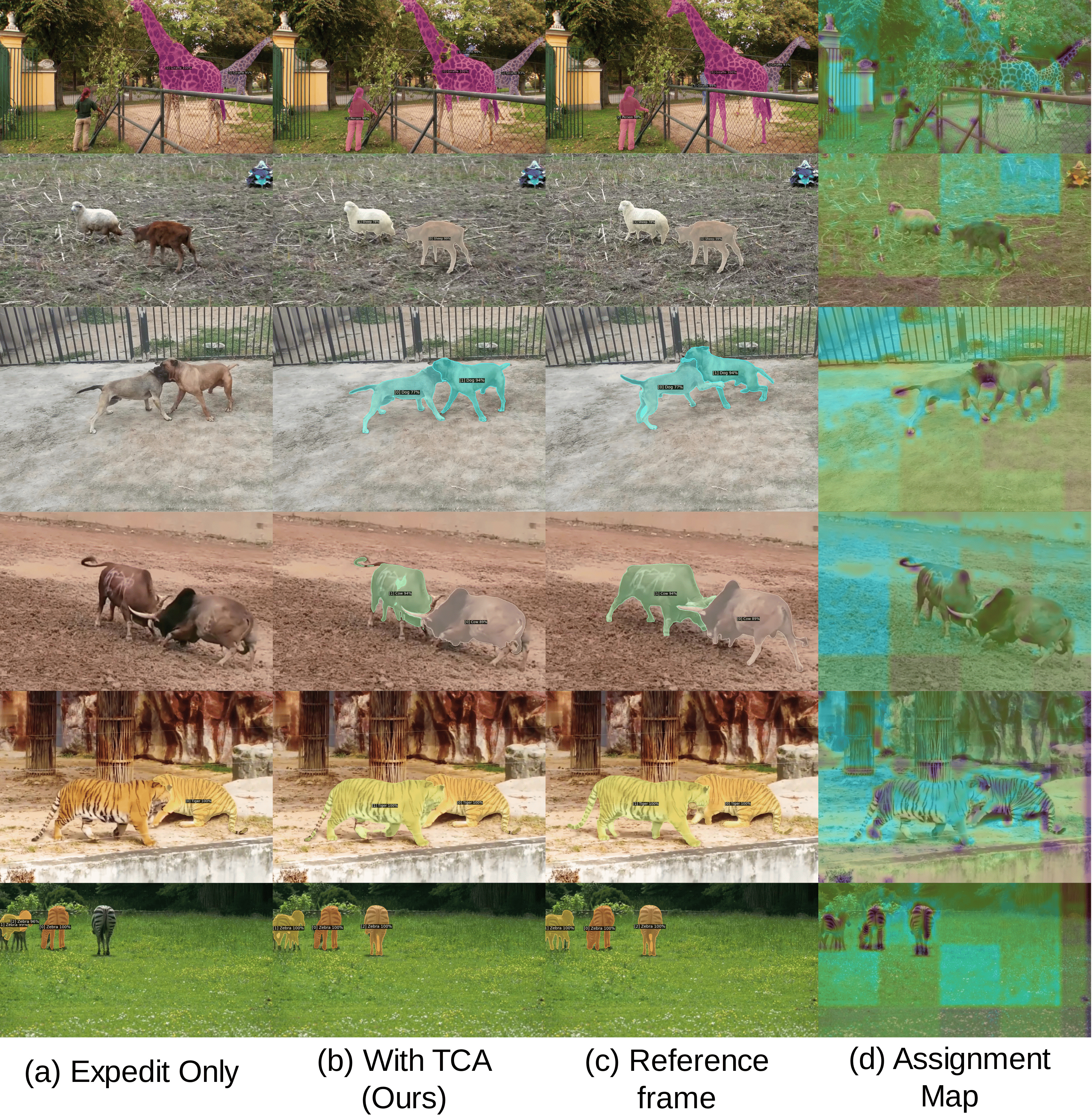}
\end{center}
   \caption{Further examples of comparisons and heatmaps from OVIS}
\label{fig:extra_ovis}
\end{figure*}
\FloatBarrier






\end{document}